\documentclass[runningheads]{llncs}
\usepackage[T1]{fontenc}
\usepackage{boldline}

\usepackage[binary-units=true, detect-all]{siunitx}[=v2]

\usepackage{amsfonts}
\usepackage{amssymb}
\usepackage{amsmath}
\usepackage{graphicx}
\usepackage{xcolor}
\usepackage{soul}
\usepackage{cite}
\usepackage{booktabs} 
\usepackage{tikz}
\usetikzlibrary{arrows.meta, positioning, calc, shapes.geometric}
\definecolor{UpColor}{HTML}{A1C6EA} 
\definecolor{LeftColor}{HTML}{F7CAC9} 
\definecolor{DownColor}{HTML}{FFDDA1} 
\definecolor{RightColor}{HTML}{A8E6CF} 

\begin{document}

\title{Anatomy Might Be All You Need: Forecasting What to Do During Surgery}

\titlerunning{Anatomy Might Be All You Need}
\author{Gary Sarwin\inst{1} \and
Alessandro Carretta\inst{2,3} \and
Victor Staartjes\inst{2}\and
Matteo Zoli\inst{3}\and
Diego Mazzatenta\inst{3}\and
Luca Regli\inst{2}\and
Carlo Serra\inst{2} \and 
Ender Konukoglu\inst{1}}
\authorrunning{G Sarwin et al.}
\institute{Computer Vision Lab, ETH Zurich, Switzerland \and
Department of Neurosurgery, University Hospital of Zurich, Zurich, Switzerland \and 
Department of Biomedical and Neuromotor Sciences (DIBINEM), University of Bologna, Bologna, Italy}
\maketitle              %
\begin{abstract}

Surgical guidance can be delivered in various ways. In neurosurgery, spatial guidance and orientation are predominantly achieved through neuronavigation systems that reference pre-operative MRI scans. Recently, there has been growing interest in providing \emph{live} guidance by analyzing video feeds from tools such as endoscopes. Existing approaches, including anatomy detection, orientation feedback, phase recognition, and visual question-answering, primarily focus on aiding surgeons in assessing the current surgical scene. This work aims to provide guidance on a finer scale, aiming to provide guidance by \emph{forecasting the trajectory of the surgical instrument}, essentially addressing the question of what to do next. To address this task, we propose a model that not only leverages the historical locations of surgical instruments but also integrates anatomical features. Importantly, our work does not rely on explicit ground truth labels for instrument trajectories. Instead, the ground truth is generated by a detection model trained to detect both anatomical structures and instruments within surgical videos of a comprehensive dataset containing pituitary surgery videos. By analyzing the interaction between anatomy and instrument movements in these videos and forecasting future instrument movements, we show that anatomical features are a valuable asset in addressing this challenging task. To the best of our knowledge, this work is the first attempt to address this task for manually operated surgeries.
\end{abstract}
\section{Introduction}

\paragraph{Surgical guidance} is one of the hallmark problems in computed-aided interventions. 
The main goal is simply to build algorithms to help the surgeon during the surgery.
This guidance can come in multiple forms, in neurosurgery the major ones being neuronavigation and surgical action or phase recognition \cite{hartl2013worldwide, orringer2012neuronavigation, garrow2021machine}. 
While the former is crucial for helping surgeons orient themselves in the anatomy, the latter can help the organization around the surgery as well as the training of new surgeons. 
Both of these guidance forms have seen tremendous advances over the last years. 
In this work, we explore and assess the possibility of a new form of guidance, forecasting surgical tool movement. While this and autonomous surgery have been explored for surgical robots where kinematics are available \cite{weede2011intelligent, qin2020davincinet, kim2024surgical, saeidi2022autonomous}, to the best of our knowledge, this is the first work addressing this task in manually operated surgeries.

This forecasting problem can be coarsely defined as predicting the next movement of the surgical tool. 
The first step towards this direction is simply predicting the next location of the surgical tool. 
In the setting of surgical videos or video streams, this corresponds to predicting the location of the surgical tool in the next video frames given the previous ones. 
This is a narrower subproblem, as well as the first step, of the broader problem of predicting the surgeon's next action. 
It can also be viewed as a narrower version of another broader problem, forecasting frames in a video. 

The applications of forecasting surgical tool movement can be multiple. 
First, it is a natural next step for surgical guidance. 
It goes one step further than providing information for self-orientation, it provides guidance towards how to move the surgical tool in a given scenario. 
A successful forecasting model trained with expert surgeons' actions can guide surgeons less experienced with a particular surgery by suggesting a surgical tool's next movements. 
Such forecasting can also be an integral component of a fully autonomous surgical robot. 

Recognition of surgical tools in surgical videos as well as recognition of surgical actions being performed in a set of observed frames have received ample attention from the research community. 
Advanced tools of today can make accurate predictions for both problems \cite{das2024pitvis, padoy2019machine}. 

Forecasting, however, is arguably a much more challenging problem than these two. 
This is akin to the more challenging nature of the next frame prediction problem compared to parsing a given scene or action recognition in a given video sequence.
Forecasting the movement of a surgical tool requires a full understanding of the scene as well as the past movements of the tool. 
Until recently, a full understanding of a surgical scene was beyond reach. 
Advances in this problem, especially through the recent works in structure recognition and detection \cite{staartjes2021machine, sarwin2023live}, have opened up new opportunities towards tackling this challenging forecasting problem. 
Here, we build upon these recent advances and, to the best of our knowledge, propose one of the first attempts to solving the forecasting problem. 

In this work, we present a deep learning model to forecast the location of a surgical tool in the next frames given previous frames of a surgical video. 
The model uses a very simple neural network architecture to predict the location of a surgical tool in the next 8 or 16 frames given the previous 64 frames. 
We evaluate the forecasting accuracy using different levels of information on the scene, including the location of the surgical tool, and locations of anatomical structures as predicted by a detector (YOLO \cite{wang2023yolov7}) trained on surgical videos. 
Experimental results suggest that the value of the anatomical structure detections may be the key towards solving the forecasting problem. 
This work is a first step and we hope it will pave the way towards accurate forecasting models and ultimately accurate guidance on the next surgical action. 
\section{Methods}
\begin{figure}
    \centering
    \includegraphics[width=1\linewidth]{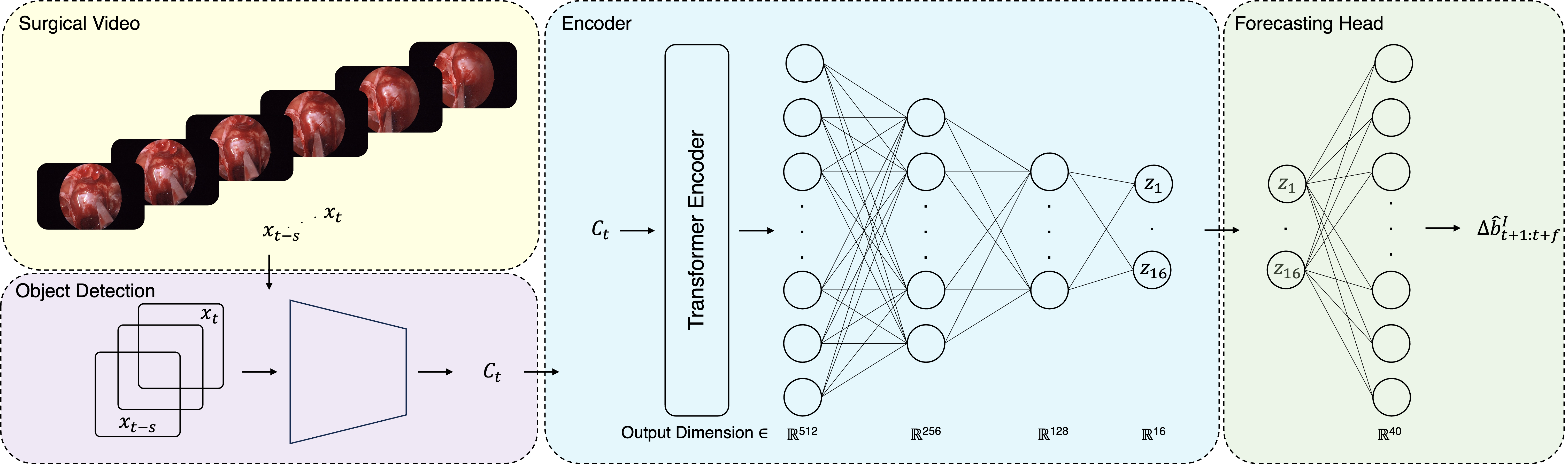}
    \caption{Overview of the proposed model pipeline. Frames extracted from a surgical video are processed through an object detection network, which identifies and localizes anatomical structures and surgical instruments across a sequence of frames. These detected sequences are then passed into an encoder to extract temporal features. Finally, the forecasting head predicts the bounding box changes of the instrument for the next $f$ frames.}
    \label{fig:model}
\end{figure}
\subsection{Problem Formulation and Approach}

Let $\mathbf{S}_{t}$ denote a sequence of endoscopic video frames $\mathbf{x}_{t-s:t}$, where $s$ is the sequence length, and $\mathbf{x}_t \in \mathbb{R}^{w \times h \times c}$ is the $t$-th frame with width $w$, height $h$, and number of channels $c$. Our primary goal is to predict the future changes in the locations of the surgical instrument bounding boxes for the future $f$ frames from $t$ to $t+f$, denoted as $\Delta \mathbf{b}_{t:t+f}^I = [\Delta \mathbf{b}_t^I, \dots, \Delta \mathbf{b}_f^I]^T \in \mathbb{R}^{f \times 4}$, where $I$ denotes the instrument and bounding boxes are represented with 4 values. Our approach involves identifying the anatomical structures and the surgical instrument in the sequence $\mathbf{S}_t$. To achieve this, object detection is performed on all frames $\mathbf{x}_{t-s:t}$ in $\mathbf{S}_{t}$, resulting in a sequence of detections $\mathbf{c}_{t-s:t}$ denoted as $\mathbf{C}_t$. A detection $\mathbf{c}_t \in \mathbb{R}^{n \times 5}$ includes binary variables $\mathbf{y}_t = [y_t^0, \dots, y_t^n] \in \{0,1\}^{n}$ indicating the presence of structures in the $t$-th frame, where $n$ is the number of structures, and bounding box coordinates $\mathbf{b}_t = [\mathbf{b}_t^0, \dots, \mathbf{b}_t^n]^T \in \mathbb{R}^{n \times 4}$. $\mathbf{C}_t$ is then used to predict the future changes of the instrument bounding boxes $\Delta \hat{\mathbf{b}}_{t+1:t+f}^I$. An overview of the proposed pipeline is visualized in Figure \ref{fig:model}.

\subsection{Object Detection}

To identify and locate anatomical structures, we utilize the YOLOv7 network in the object detection phase of our pipeline \cite{wang2023yolov7}. The detection network is trained on endoscopic videos from a training set, where frames are sparsely labeled with bounding boxes. Subsequently, the trained network is applied to all the frames of an incoming sequence and the bounding box detections are performed and form $\mathbf{C}_t$. Then $\mathbf{C}_t$ is passed as input to the forecasting network, which predicts the future instrument bounding box location and is described next. The bounding boxes are parameterized by four values: the center coordinates (\(x, y\)) and the width and height (\(w, h\)). Each value is normalized relative to the image dimensions.

\subsection{Future Instrument Location Prediction}

To learn the future bounding box locations, a forecasting model takes as input $\mathbf{C}_t$ and predicts $\Delta \hat{\mathbf{b}}_{t+1:t+f}^I$. This model's parameters are updated to ensure that the predicted future changes in surgical instrument bounding boxes $\Delta \hat{\mathbf{b}}_{t+1:t+f}^I$ fit the ground truth changes $\Delta \mathbf{b}_{t+1:t+f}^I$ in a training set. This leads to the objective to minimize for the $t$-th frame in the $m$-th training video:
\begin{equation*}
\begin{aligned}
\mathcal{L}_{m, t} = & \sum_{r=t+1}^{t+f} \left|\Delta \mathbf{b}_{m, r}^I - \Delta \hat{\mathbf{b}}_{m, r}^I\right|
 + \lambda \cdot \Bigg(1 - \frac{\langle \bar{\mathbf{v}}_{p, m, t:t+f}, \bar{\mathbf{v}}_{g, m, t:t+f} \rangle}{\|\bar{\mathbf{v}}_{p, m, t:t+f}\| \|\bar{\mathbf{v}}_{g, m, t:t+f}\|}\Bigg),
\end{aligned}
\end{equation*}
where:
\begin{itemize}
    \item $\Delta \mathbf{b}_{m, r}^I = \mathbf{b}_{m, r}^I - \mathbf{b}_{m, r-1}^I$ represents the ground truth change in bounding box values between frame $r-1$ and $r$ for the instrument I,
    \item $\Delta \hat{\mathbf{b}}_{m, r}^I = \hat{\mathbf{b}}_{m, r}^I - \hat{\mathbf{b}}_{m, r-1}^I$ represents the predicted change in bounding box values between frame $r-1$ and $r$ for the instrument I,
    \item $\bar{\mathbf{v}}_{p, m, t:t+f}$ and $\bar{\mathbf{v}}_{g, m, t:t+f}$ are the average direction vectors for the predicted and ground truth changes in x and y coordinates of the instrument bounding box over the sequence of frames $t+1:t+f$,
    \item $\langle \cdot, \cdot \rangle$ denotes the dot product,
    \item $\|\cdot\|$ represents the vector norm,
    \item $\lambda$ is a weighting factor that balances the contribution of the direction loss,
    \item $|\cdot|$ denotes the $L_1$-loss.
\end{itemize}

The first term computes the $L_1$ difference between the predicted and ground truth frame-to-frame changes in bounding box values, ensuring that the predicted changes align with the ground truth. The second term introduces a direction loss that penalizes deviations in the alignment of the predicted and ground truth trajectories by minimizing the cosine similarity between their average direction vectors. The weighting factor $\lambda$ controls the relative importance of the direction loss. The total training loss is then defined as the sum of $\mathcal{L}_{m, t}$ over all frames $t$ and training videos $m$. 

In order to train the forecasting network, we use a set of training videos separate from the videos used to train the detection network. This set is first passed through the detection network to extract bounding box predictions $\mathbf{C}_t$. The resulting detections constitute the inputs of the training samples for the forecasting network. 
The labels for these videos are extracted automatically using the detection network as well. 
For each sequence $\mathbf{S}_{m,t}$, the corresponding surgical instrument locations in the following $f$ frames, i.e., $\mathbf{b}_{m,t+1:t+f}^I$, are determined using the detection network on the corresponding frames, i.e., detection network applied on $\mathbf{x}_{t+1:t+f}$. 
Note that these detections are only used to generate the labels for training the forecasting network. At inference time, the model does not see the frames $\mathbf{x}_{t+1:t+f}$ while predicting $\Delta \hat{\mathbf{b}}_{m, r}^I$.

\section{Experiments and Results}
\subsection{Dataset}

The medical dataset used for consists of 169 videos, each recorded during a pituitary surgery (transsphenoidal adenomectomy) from a unique patient. These videos, collected over a 10-year period using various endoscopes and sourced from multiple centers, were made available under general research consent. Expert neurosurgeons provided annotations, identifying 16 classes in total: 15 representing distinct anatomical structures and one representing surgical instruments. In total, the dataset includes approximately 19,000 labeled frames. Generally, there is one instance of each anatomical class per video, while multiple different instruments are grouped into a single instrument class. Of the 169 videos, 77 were allocated for training and validation of the object detection model, 75 were used for training and validation of the model for surgical action forecasting, and the remaining 17 for testing. Although data originates from multiple sites, potential biases may still arise due to the proximity of the sites.

\subsection{Implementation Details}

The YOLO network was trained with identical parameters and implementation as reported in \cite{wang2023yolov7, sarwin2023live}.

The forecasting model integrates a transformer encoder comprising six transformer encoder layers, each with five attention heads. The input to the transformer encoder has a size of $s \times (n+1) \times 5$, where $s$ represents the sequence length, set to 64 frames, and $n+1$ represents the 15 anatomical classes plus the additional surgical instrument class. Sinusoidal positional encodings are applied to these inputs to retain the temporal dimension of the sequence, and the transformer encoder output is fed through a series of three fully connected layers with output dimensions of 512, 256, and 128, respectively, using ReLU activation functions between the layers. 
The final fully connected layer outputs a latent representation $\mathbf{z}_t \in \mathbb{R}^{16}$ with 16 dimensions.

The decoder is implemented as a single linear layer, designed to take the latent vector $\mathbf{z}_t$ as input. This vector is processed through the layer to reach the final output of $\text{predicted sequence length} \times 4$, where each predicted vector describes the bounding box changes.

The output consists of frame-to-frame changes in bounding box coordinates, rather than absolute bounding box positions, to enable the model to effectively predict motion dynamics over the sequence. The model also incorporates cosine similarity-based direction loss to ensure directional consistency in predicted trajectories.

For training, the AdamW optimizer \cite{loshchilov2017decoupled} is employed alongside a warm-up scheduler, which linearly increases the learning rate from $0$ to \num{1e-4} over 60 warm-up epochs, and $\lambda$ was set to 0.5. The model is trained for a total of 75 epochs, and 150 for the task of predicting the instrument trajectory in the next 8 and 16 frames, respectively. 
\subsection{Results}
\paragraph{Evaluation metric:} Considering the challenging nature of this task, we assess the forecasting performance through a classification metric. More specifically, we  classify the predicted direction of the instrument movement into four principal directions 
: \textit{up}, \textit{down}, \textit{left}, and \textit{right}. The classification is based on the angular range \(\theta\) of the movement, defined in degrees as follows:
\begin{center}
\begin{tikzpicture}[scale=0.7] 
    \draw[thick] (0,0) circle (2cm);

    \fill[UpColor] (0,0) -- ({2*cos(45)},{2*sin(45)}) arc[start angle=45, end angle=135, radius=2cm] -- cycle; 
    \fill[LeftColor] (0,0) -- ({2*cos(135)},{2*sin(135)}) arc[start angle=135, end angle=225, radius=2cm] -- cycle;
    \fill[DownColor] (0,0) -- ({2*cos(225)},{2*sin(225)}) arc[start angle=225, end angle=315, radius=2cm] -- cycle;
    \fill[RightColor] (0,0) -- ({2*cos(315)},{2*sin(315)}) arc[start angle=315, end angle=405, radius=2cm] -- cycle;

    \node[above] at (0,2.5) {\textbf{Up}};
    \node[below] at (0,-2.5) {\textbf{Down}};
    \node[left] at (-2.5,0) {\textbf{Left}};
    \node[right] at (2.5,0) {\textbf{Right}};

    \draw[dashed, thick] (0,0) -- (1.414,1.414); 
    \draw[dashed, thick] (0,0) -- (-1.414,1.414); 
    \draw[dashed, thick] (0,0) -- (-1.414,-1.414); 
    \draw[dashed, thick] (0,0) -- (1.414,-1.414); 

    \node[align=center, font=\small] at ($(1.414,1.414) + (0.4,0.4)$) {\(45^\circ\)};
    \node[align=center, font=\small] at ($(-1.414,1.414) + (-0.4,0.4)$) {\(135^\circ\)};
    \node[align=center, font=\small] at ($(-1.414,-1.414) + (-0.4,-0.4)$) {\(225^\circ\)};
    \node[align=center, font=\small] at ($(1.414,-1.414) + (0.4,-0.4)$) {\(315^\circ\)};
\end{tikzpicture}
\end{center}

\noindent Here, \(\theta\) represents the predicted angle of movement measured in degrees. 
\paragraph{Accounting for noise in ground truth movement:} 
The ground truth motion of the surgical tools was estimated using their bounding box locations predicted with the detection network using the frames as input. 
These predictions contained jitter noise and this induces noise in the ground truth movements for the surgical tools. 
While this noise is desirable for training, since it adds to the robustness of the tool, it is detrimental to an accurate evaluation of the performance. 
We therefore additionally evaluated the performance of the tool for movements larger than a certain threshold, which are unlikely due to the jitter of the predictions of the detection network. 
To filter out noisy movements, we test movements of which the magnitude is greater than $0.1$ and $0.05$, which when considering a horizontal movement and an image size of $1920\times1280$, corresponds to 192 and 96 pixels respectively. The filtering resulted in around $40000$ and $77000$ test samples for the case of predicting 8 frames, respectively. These samples were then also used to generate the results of the models that predict 16 frames. 
\paragraph{Quantitative results:} 
\begin{table}[t!]
\centering
\begin{tabular}{@{}lcccccc@{}}
\toprule
\textbf{Model} & \multicolumn{6}{c}{\textbf{\# Frames $f$}}  \\
\cmidrule(lr){2-7} 
& \multicolumn{3}{c}{\textbf{$8$}} & \multicolumn{3}{c}{\textbf{$16$}} \\
\cmidrule(lr){2-4} \cmidrule(lr){5-7} 
& \textbf{$0.1$} & \textbf{$0.05$} & \textbf{$0$} & \textbf{$0.1$} & \textbf{$0.05$} & \textbf{$0$} \\
\midrule
Anatomy + Instrument & \textbf{60.58} & \textbf{51.84} & \textbf{38.04} & \textbf{55.86} & \textbf{49.81}& \textbf{40.34}\\
Instrument & 50.71 & 42.51 & 33.19 &52.17 & 46.33 & 37.9\\
Random & 24.71 & 24.98 & 25.02 & 25.25 & 25.17& 25.09\\
\bottomrule
\end{tabular}\caption{Ablation study on direction classification accuracy (in percentage). The results are presented for two scenarios: predicting the next 8 frames and predicting the next 16 frames. Additionally, the performance is evaluated across various threshold magnitudes of movement.}
\label{tab:direction_accuracy_results}
\end{table}

The results, summarized in Table~\ref{tab:direction_accuracy_results}, demonstrate a significant improvement in direction classification accuracy when anatomical detections are incorporated into the forecasting model. For instance, with a movement threshold of $0.1$, the model achieves 60.58\% accuracy for 8-frame predictions when both anatomy and instrument detections are used, compared to 50.71\% when only instrument detections are used. Similarly, for 16-frame predictions, the accuracy is 55.86\% with anatomy and instrument detections versus 52.17\% with only instrument detections. As expected, the accuracy decreases when including all movements and no thresholding is performed.

\paragraph{Qualitative Results:} 
\begin{figure}[h!]
    \centering
    \includegraphics[width=1\linewidth]{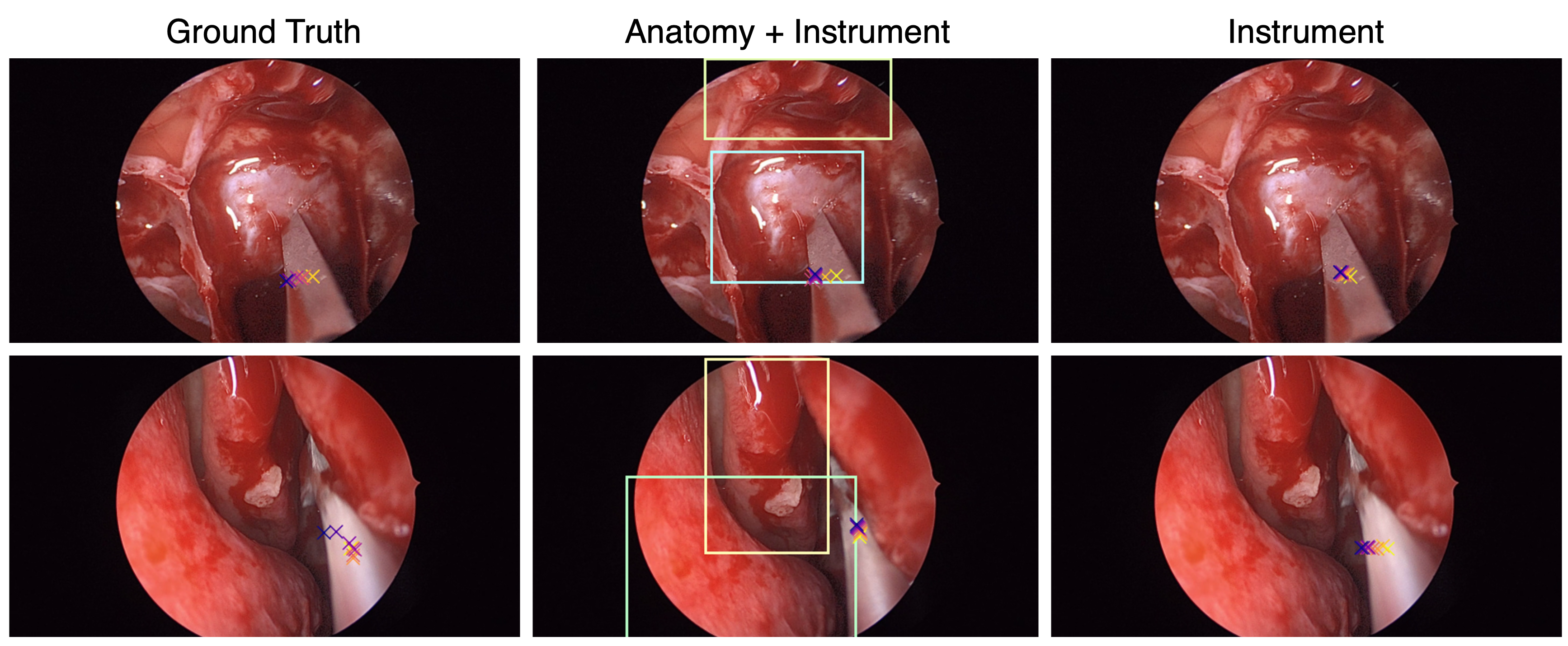}\caption{Qualitative comparison of the forecasting models predicting 8 frames. For visualization purposes, only the predicted future bounding box centers are displayed. The yellow crosses indicate predictions closest to the current frame, while the purple crosses represent predictions furthest into the future. Additionally, the anatomy detections for the model that takes these as input are visualized.}
    \label{fig:qualitative}
\end{figure}
To further understand the impact of incorporating anatomy detections into the forecasting pipeline, qualitative results are presented in Figure~\ref{fig:qualitative}. The predicted trajectories for the surgical instrument are shown alongside the ground truth trajectories.
\section{Conclusion}

In this work, we have proposed a novel approach to forecasting surgical tool movements using endoscopic videos, leveraging anatomical structure detection and a transformer-based forecasting model. Our results demonstrate that incorporating anatomical context significantly improves the predictive performance of surgical tool trajectory forecasting.

This study represents a foundational exploration of forecasting surgical tool movements. The ability to anticipate tool movements offers a range of potential applications, from providing real-time decision support for surgeons to facilitating the development of surgical robots. Moreover, this forecasting framework could serve as a critical component in surgical training, offering insights into expert movement patterns and enabling feedback for trainees.

\subsection{Limitations and Future Work}
\label{limitations}
While promising, there are various limitations to this method. As of now, the model treats all surgical instruments as identical which hinders the understanding of surgical actions as each instrument has a specific purpose. Additionally, the framework only predicts the actions of one surgical instrument, which can create confusion for the model as soon as there are two instruments in view. Therefore, we suspect improvements if the instruments are classified and the model can handle multiple instruments as input. Additionally, we expect further improvements by incorporating an autoregressive architecture and by analyzing visual features with the forecasting model to enable more accurate predictions. These enhancements are part of our planned future work.

\begin{credits}
\subsubsection{\ackname} This work funded by the SNSF (Project IZKSZ3\_218786).

\subsubsection{\discintname}
The authors declare no conflict of interest.
\end{credits}
\bibliographystyle{splncs04}
\bibliography{references}
\end{document}